# DynSegNet:Dynamic Architecture Adjustment for Adversarial Learning in Segmenting Hemorrhagic Lesions from Fundus Images

**Zesheng Li[1], Minwen Liao[1], Haoran Chen[2], Yan Su[1], Chengchang Pan[1*]**, and **Honggang Qi[1*]**

[1]School of Computer Science and Technology, University of Chinese Academy of Sciences
[2]School of Information Engineering, Sichuan Agricultural University
lizesheng@stu.sufe.edu.cn

**Abstract**

The hemorrhagic lesion segmentation plays a critical role in ophthalmic diagnosis, directly influencing early disease detection, treatment planning, and therapeutic efficacy evaluation. However, the task faces significant challenges due to lesion morphological variability, indistinct boundaries, and low contrast with background tissues. To improve diagnostic accuracy and treatment outcomes, developing advanced segmentation techniques remains imperative. This paper proposes an adversarial learning-based dynamic architecture adjustment approach that integrates hierarchical U-shaped encoder-decoder, residual blocks, attention mechanisms, and ASPP modules. By dynamically optimizing feature fusion, our method enhances segmentation performance. Experimental results demonstrate a Dice coefficient of 0.6802, IoU of 0.5602, Recall of 0.766, Precision of 0.6525, and Accuracy of 0.9955, effectively addressing the challenges in fundus image hemorrhage segmentation.

## 1 Introduction

Fundus hemorrhage serves as a critical pathological marker for ocular diseases such as diabetic retinopathy, where precise segmentation holds vital clinical significance for early diagnosis and treatment planning. By accurately identifying the location and extent of hemorrhagic lesions, clinicians can develop more effective therapeutic strategies, thereby improving patient outcomes and prognosis.

_______________

\* Corresponding author.

However, despite significant advancements in medical image segmentation, traditional methods like Fully Convolutional Networks (FCN) and DeepLabV3+ still face formidable challenges in processing complex hemorrhagic lesions in fundus images.

Atlas et al. (2017) demonstrated that traditional threshold-based methods (e.g., ANFIS combined with particle swarm optimization) could partially segment regular lesions but exhibited significantly reduced sensitivity for polymorphic, low-contrast hemorrhagic regions. These lesions are characterized by morphological diversity, blurred boundaries, and low contrast with surrounding tissues. Wang S. et al. (2019) found that boundary ambiguity could reduce segmentation Dice coefficients by up to 12.7%, rendering accurate segmentation exceptionally challenging. Furthermore, the substantial variability in lesion size and morphology necessitates effective fusion of multi-scale features to comprehensively capture pathological information. For instance, Khojasteh P. et al. (2018) demonstrated that a ten-layer CNN architecture with probabilistic multi-scale feature fusion significantly improved hemorrhage detection rates, though its computational complexity hindered clinical adoption.

To address these challenges, this study proposes a novel fundus hemorrhage segmentation framework integrating adversarial learning with dynamic architectural adaptation. The framework leverages a generator-discriminator interaction mechanism: the generator performs feature extraction and preliminary segmentation, while the discriminator evaluates and refines the segmentation outputs. The generator employs a hierarchical U-shaped encoder-decoder enhanced with residual blocks to strengthen feature representation capabilities. Additionally, it incorporates spatial attention mechanisms and an Atrous Spatial Pyramid Pooling (ASPP) module to capture multi-scale contextual information.

Our contributions are threefold: (1) We design a generator network that enables autonomous focusing on subtle textural features of hemorrhagic lesions. (2) We develop a discriminator network based on PatchGAN and dynamic adjustment mechanisms to better capture fine-grained details. (3) Our model demonstrate exceptional performance across key evaluation metrics, achieving a Dice coefficient of 0.6802, IoU of 0.5602, Recall of 0.766, Precision of 0.6525

## 2 Related Work

### 2.1 Fundamentals of Deep Learning

The application of deep learning in medical image analysis has become extremely widespread. With continuous technological advancements and algorithm optimization, deep learning techniques have demonstrated increasingly significant roles in medical image recognition, classification, segmentation, and disease diagnosis, leveraging their powerful data processing and pattern recognition capabilities. A prime example is the U-Net segmentation network, an innovative architecture proposed by Ronneberger et al. (2015). By constructing a U-shaped network structure, it skillfully integrates semantic and positional information. The encoder-decoder architecture, combined with skip connections, enables effective concatenation of features from the downsampling process with those in the decoder phase, thereby preserving finer details in pixel-level classification tasks. This design has made U-Net particularly outstanding in medical image segmentation, especially in retinal vessel segmentation.

Building upon this foundation, researchers have conducted numerous valuable explorations based on CNNs and U-Net. For instance, Uysal et al. (2020) implemented a fully convolutional neural network to segment vessels from grayscale fundus images. Gu et al. (2019) proposed a Contextual Encoder Network (CENET), achieving high-precision 2D vessel segmentation by constraining complex high-level information while retaining spatial details. Hu et al. (2018) combined CNNs with fully connected Conditional Random Fields (CRFs) to accurately segment vessels from color fundus images, employing a multi-scale CNN architecture and an enhanced cross-entropy loss function to drive robust training. Additionally, Shin et al. (2019) innovatively bridged CNN architectures with Graph Neural Networks (GNNs), proposing a Vascular Graph Network (VGN), which opened new avenues for retinal vessel segmentation.

### 2.2 GANs and Their Variants

In recent years, Generative Adversarial Networks (GANs) have sparked a surge of interest in medical image analysis due to their unique adversarial mechanism. Emami et al. (2018) leveraged GANs to synthesize CT images from magnetic resonance imaging (MRI), demonstrating their immense potential for cross-modal image generation. In the field of medical image segmentation, GANs and their variants have become a research hotspot through their integration with U-Net architectures.

By leveraging the adversarial interplay between generators and discriminators, GANs excel at producing high-quality segmentation results, particularly in few-shot learning scenarios and data augmentation. For instance, Wu et al. (2019) proposed U-GAN, which ingeniously integrated U-Net with attention mechanisms to enhance retinal vessel segmentation accuracy and noise robustness through adversarial training. Meanwhile, Xue et al. (2017) introduced SegAN, employing multi-scale L1 loss functions to improve the stability and precision of GANs in medical image segmentation, outperforming traditional U-Net methods.

Researchers have further explored GANs for specialized segmentation tasks. Guo et al. (2020) achieved remarkable accuracy in pectoralis major segmentation from mammograms by decoupling boundary detection and shape prediction, particularly excelling in handling ambiguous boundaries. Fan et al. (2023) developed U-Patch GAN, combining U-Net with PatchGAN to fuse multimodal brain images via self-supervised learning. This approach not only enhanced fusion quality but also preserved functional and structural details, providing more reliable data for clinical diagnosis.

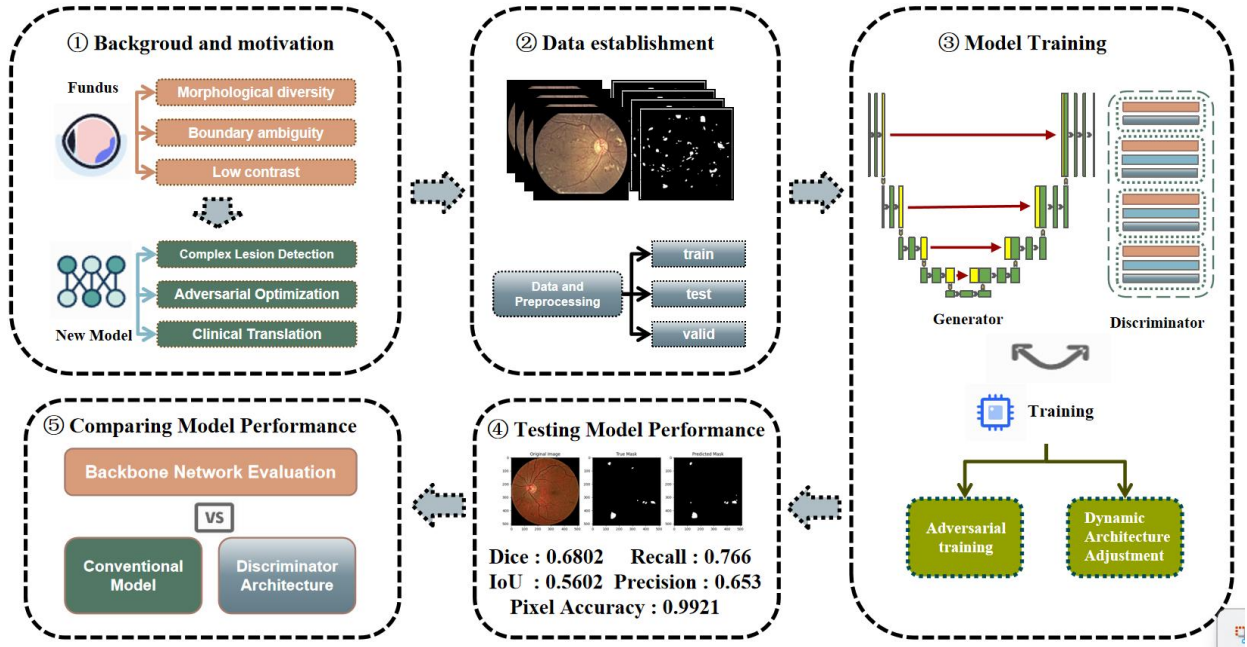

Figure 1: From fundus image preprocessing to clinical validation: Adversarial-driven dynamic with ASPP multi-scale attention resolves lesion diversity/boundary ambiguity (Dice:0.6802).

Efforts to address domain-specific challenges have also flourished. Lei et al. (2020) designed UNet-SCDC-GAN, incorporating dense convolutional U-Net and dual discriminators to simultaneously optimize boundary delineation and contextual coherence, significantly improving skin lesion segmentation. Sun et al. (2020) proposed MM-GAN for 3D MRI augmentation, mitigating data scarcity in tumor segmentation while anonymizing patient data to safeguard privacy. Additionally, Shi et al. (2020) introduced AUGAN, which synergized U-Net with deep aggregation modules to automate nodule segmentation in chest CT scans, achieving state-of-the-art precision and Hausdorff distance metrics.

## 2.3 Advances in Adversarial Learning for Retinal Image Segmentation

In the specialized field of retinal fundus image analysis, the integration of GANs with U-Net architectures has demonstrated remarkable innovative potential. Lin et al. (2021) enhanced U-Net's performance in retinal vessel detail segmentation by incorporating bidirectional ConvLSTM and GANs, achieving significant improvements in detecting microvessels and terminal branches. Meanwhile, Guo et al. (2020) synergized Dense U-Net with GANs, enabling efficient automated retinal vessel segmentation through multi-scale feature extraction and adversarial training, outperforming numerous existing methods.

A groundbreaking advancement came from Pachade et al. (2021) with their NENet framework. By combining Nested EfficientNet with adversarial learning, this approach realized joint optic disc and cup segmentation using multi-scale analysis and attention mechanisms. The model not only surpassed state-of-the-art methods but also exhibited exceptional generalization capabilities, making it particularly suitable for automated glaucoma screening systems. Further pushing the boundaries, Kar et al. (2023) integrated multi-scale residual convolutional networks with GANs. Through optimized hybrid loss functions, their method achieved benchmark-setting accuracy in retinal vessel segmentation, establishing new performance standards in the field.

## 3 Methodology

This study proposes a segmentation framework based on adversarial learning and dynamic architectural adaptation (as illustrated in Figure 2). The framework comprises three core components: a Generator network, a Discriminator network, and an embedded dynamic adaptation module, which collaboratively achieves precise lesion segmentation through multi-phase optimization

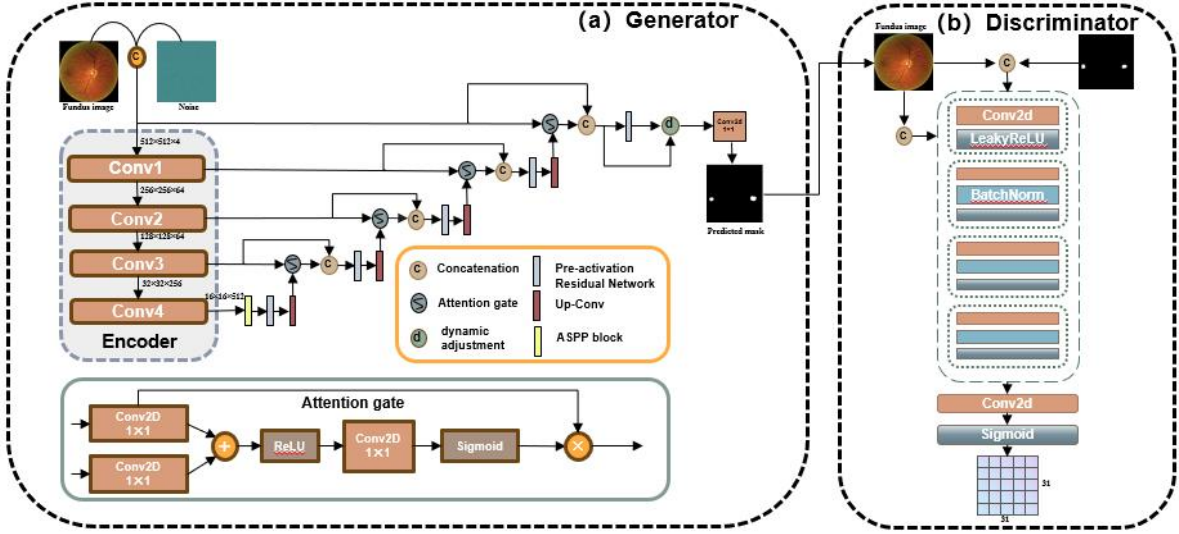

Figure 2: DynSegNet's GAN architecture: (a) generator with hierarchical U-shaped encoder-decoder (b) PatchGAN discriminator

## 3.1 Generator Design

The generator, as the core component of the segmentation framework, achieves precise localization and feature representation of lesion regions through multi-module collaboration. The overall architecture employs an improved Hierarchical U-shaped encoder-decoder as the backbone, integrates residual modules to enhance gradient propagation, incorporates a spatial attention mechanism to focus on critical areas, and introduces an Atrous Spatial Pyramid Pooling (ASPP) module to capture multi-scale contextual information. Through synergistic interactions among these modules, the framework effectively addresses challenges in medical segmentation such as diverse lesion morphology and blurred boundaries.

### 3.1.1 Attention Mechanism

The attention mechanism is a technique that mimics the selective perception of human vision. Its core concept lies in automatically focusing on regions most critical to the current task (e.g., image segmentation) by calculating the importance weights of different regions or features, while suppressing the influence of irrelevant or less relevant areas.

Specifically, the mechanism achieves this by introducing attention coefficients $\alpha_i$. These coefficients, ranging between 0 and 1, quantify the relative importance of different regions/features for the task. When $\alpha_i$ approaches 1, it indicates high relevance of the corresponding region, prompting the model to allocate stronger attention; conversely, values near 0 suggest low importance, leading to attenuated focus.

The output of the attention mechanism is a weighted feature representation that intensifies focus on task-critical regions/features, thereby enhancing the model's performance and accuracy. Mathematically, this operation can be formulated as the element-wise multiplication between the input feature map and attention coefficients:

$$\hat{x}^l_{i,c} = \{x^l_{i,c}, \alpha^l_i\}_{i=1}^n$$

Where, $x^l = \{x^l_{i,c}\}_{i=1}^n$ is the feature map for pixel $i$ in layer $l$ and class $c$. The specific implementation of the attention gate architecture referenced here is illustrated in Figure 2.

### 3.1.2 Atrous Spatial Pyramid Pooling

Atrous Spatial Pyramid Pooling (ASPP) is a technique in deep learning designed to capture multi-scale contextual information. By integrating multiple parallel dilated convolutions with varying dilation rates, ASPP expands the receptive field of convolutional kernels without increasing computational complexity, thereby effectively capturing broader contextual information. The core principle of dilated convolution lies in adjusting the dilation rate of the kernel, enabling the model to simultaneously extract features at different scales. This capability is particularly critical for processing medical images characterized by variable lesion scales and intricate structural patterns.

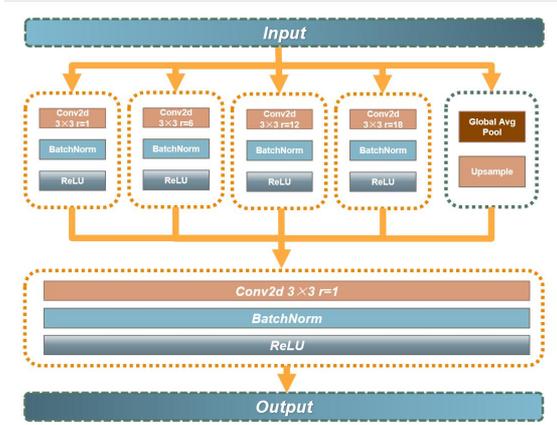

Figure 3: Atrous spatial pyramid pooling block.

In our implementation, the ASPP module comprises five parallel branches: four dilated convolutional branches and one global average pooling branch. As illustrated in Figure 3, each dilated convolutional branch employs distinct dilation rates (e.g., 1, 6, 12, 18) to capture features at varying scales. The global average pooling branch compresses the entire feature map into a global context vector, encapsulating holistic semantic information.

The outputs from all branches are concatenated along the channel dimension, forming a multi-scale fused representation. This concatenated feature tensor is subsequently processed through a 1×1 convolutional layer for dimensionality integration, ultimately generating the refined output feature map. This hierarchical pooling strategy enables the model to adaptively handle input size variations while preserving spatial-semantic coherence, thereby significantly enhancing the model's generalization capability across diverse medical imaging modalities.

## 3.2 Discriminator

In Generative Adversarial Networks (GANs), the design of the discriminator plays a pivotal role in determining model performance and training stability. The discriminator's primary objective is to distinguish real samples from generated ones, thus providing the generator with effective gradient signals for optimization. In this study, we adopt an enhanced PatchGAN architecture as the discriminator backbone, integrated with a dynamic spectral normalization framework to further improve model robustness and convergence stability.

### 3.2.1 PatchGAN Architecture

The core advantage of PatchGAN lies in its ability to capture localized patterns and textural details within images. Specifically, PatchGAN operates through a sliding window mechanism that partitions the input image into multiple fixed-size patches. Each patch undergoes independent classification, with the final discrimination score obtained by averaging all patch-level predictions through global average pooling. This architectural design enables two critical benefits for high-resolution medical image processing: (1) Maintains fine-grained structural features (e.g., lesion boundaries, tissue textures) through patch-wise analysis. (2) Reduces complexity from $O(n^2)$ to $O(n)$ through parameter-sharing convolutional implementation

The PatchGAN loss can be expressed as:

$$L_{PatchGAN} = -\frac{1}{N}\sum_{i=1}^{N}[y_i \log(D(x_i)) + (1-y_i)\log(1-D(x_i))]$$

Where, N presents the number of patches, $y_i$ is the true label(0 or 1), and $D(x_i)$ is the output of the discriminator for the i-th patch.

### 3.2.2 Dynamic Adaptation Framework

To further enhance model performance, we integrate a dynamically adaptive network architecture into the discriminator. This framework continuously regulates the weights of generator components and residual blocks through real-time adversarial loss feedback, thereby optimizing feature extraction capabilities.

The dynamic adaptation module is optimized via backpropagation, with its weight update rule formulated as:

$$\theta_{new} = \theta_{old} - \alpha \cdot \nabla_\theta L_{adv}$$

Here, $\theta_{new}$ denotes the updated weights, $\theta_{old}$ represents the pre-updated weights, $\alpha$ is the learning rate, and $\nabla_\theta L_{adv}$ indicates the gradient of the adversarial loss $L_{adv}$ with respect to the weights.

In this study, the learning rate η is set to 0.01. The dynamic adaptation module stabilizes the adversarial training between the generator and discriminator, enabling the generator to more effectively learn multi-scale features, thereby improving the overall performance of the model.

## 4 Experiment

This section presents the experimental results of our model conducted on multiple public datasets. The results demonstrate that our model exhibits outstanding performance in fundus image lesion segmentation tasks, achieving precise segmentation of pathological regions. This achievement provides strong and reliable support for medical diagnosis, contributing to improved diagnostic accuracy and efficiency.

### 4.1 Datasets and Preprocessing

This study employs four publicly available fundus image datasets for experimental validation, with specific information as follows: (1) IDRiD dataset (Porwal et al., 2018), containing fundus images from 79 patients with diabetic retinopathy; (2) DDR dataset (Li et al., 2019), comprising 757 samples of diabetic retinopathy at different severity levels; (3) Retinal-lesions dataset (Wei et al., 2020), providing 1,302 images with circular approximation annotations for hemorrhagic regions ; (4) FGADR dataset (Zhou et al., 2021), including 1,842 high-resolution fundus images with pixel-level annotations. These four datasets exhibit significant differences in sample size, lesion types, annotation granularity, and image quality, establishing a multi-dimensional validation benchmark for assessing the model's generalization capability under divergent labeling protocols.

| Name | Number | Characteristics |
| --- | --- | --- |
| IDRiD | 79 | Fundus images from patients with diabetic retinopathy |
| DDR | 757 | Diabetic retinopathy samples at different severity levels |
| Retinal-lesions | 1,302 | Images covering multiple retinal pathology features |
| FGADR | 1,842 | fundus images with pixel-level annotations |

Table 1 Fundus Image Datasets for Experimental Validation

During the data preprocessing stage, a stratified random sampling strategy was adopted to divide each dataset into training, test, and validation sets at a 7:2:1 ratio, ensuring balanced distribution of each category. All input images were uniformly resampled to 512×512 pixel resolution using a bilinear interpolation algorithm, and standardized processing (Normalization) was applied to linearly map pixel values to the [0,1] range. This preprocessing pipeline effectively eliminated domain shift issues caused by device heterogeneity while improving model training efficiency through reduced computational redundancy. To validate the effectiveness of data augmentation, dynamic Gaussian noise injection was implemented during the training phase to enhance data augmentation and expand the diversity of training samples.

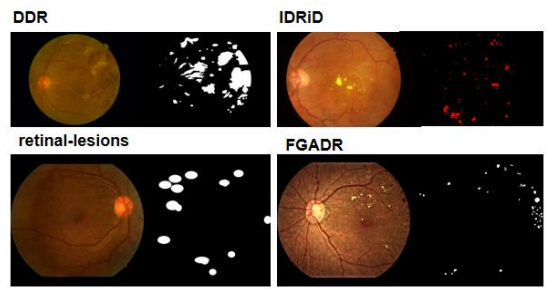

Figure 4: Sample fundus images with corresponding ground-truth masks from the datasets used in this study.

### 4.2 Evaluation Metrics

In semantic segmentation tasks, to comprehensively and accurately measure the performance of a model, multiple evaluation metrics are typically employed. These metrics reflect the model's segmentation effect on the target from different perspectives. The Dice coefficient is a commonly used statistical metric for measuring the similarity between two samples. In semantic segmentation, it assesses the model's performance by calculating the ratio of the intersection to the union between the predicted segmentation result and the true annotation. The ser its value is to 1, the more consistent the model's segmentation result is with the actual situation. The calculation formula for the Dice coefficient is:

$$Dice = \frac{2 \times |P \cap G|}{|P| + |G|}$$

| Experiment | Detail | Dice | IoU | Precision | Recall | PA |
|---|---|---|---|---|---|---|
| | ResNet-50 | 0.6820 | 0.5602 | 0.7660 | 0.6525 | 0.9955 |
| | ResNet-34 | 0.6251 | 0.4894 | 0.6784 | 0.6263 | 0.9933 |
| | ResNet-18 | 0.6400 | 0.5120 | 0.724 | 0.6190 | 0.9943 |
| | VGG | 0.6640 | 0.5395 | 0.7474 | 0.6640 | 0.9949 |
| Backbone Network | MobileNet | 0.5800 | 0.4522 | 0.6968 | 0.5605 | 0.9886 |
| | AlexNet | 0.6629 | 0.5409 | 0.7801 | 0.6136 | 0.9921 |
| | GoogLeNet | 0.6324 | 0.5033 | 0.7263 | 0.5970 | 0.9916 |
| | DenseNet | 0.6506 | 0.5203 | 0.7507 | 0.6094 | 0.9913 |
| | ViT | 0.5400 | 0.4150 | 0.7762 | 0.4619 | 0.9901 |
| | ResNeXt | 0.5455 | 0.4147 | 0.7764 | 0.4619 | 0.9901 |

Table 2 Comparative Results of Backone Network Replacement

Where, $P$ represents the set of pixels in the predicted segmentation result, $G$ represents the set of pixels in the ground truth annotation, $|P \cap G|$ indicates the number of pixels in the intersection of the predicted result and the ground truth annotation, $|P|$ and $|G|$ respectively represent the total number of pixels in the predicted result and the ground truth annotation.

The Intersection over Union (IoU) represents the ratio of the number of pixels in the intersection of the predicted segmentation area and the true segmentation area to the number of pixels in their union. A higher IoU value indicates that the model is more accurate in localizing and capturing the shape of the target during the segmentation process, and can better reflect the model's coverage of the target area. The formula for calculating IoU is as follows:

$$IoU = \frac{|P \cap G|}{|P \cup G|}$$

Where, $|P \cup G|$ represents the number of pixels in the union of the predicted results and the ground truth annotations.

The F1 Score is the harmonic mean of Precision and Recall, which comprehensively considers the model's ability to correctly identify positive samples and the proportion of positive samples that are correctly classified as such. The F1 Score strikes a balance between Precision and Recall, providing a comprehensive measure of the overall performance of the model. The formula for calculating the F1 Score is:

$$F1 = 2 \times \frac{Precision \times Recall}{Precision + Recall}$$

Precision represents the proportion of actual positive samples among the results predicted as positive by the model. It focuses on the accuracy of the model's predictions, aiming to avoid excessive false positives. The formula for calculating Precision is:

$$Precision = \frac{TP}{TP + FP}$$

Recall reflects the proportion of actual positive samples that are correctly predicted as positive by the model among all actual positive samples. It emphasizes the model's ability to comprehensively identify positive samples, ensuring that as many true target regions as possible are detected. The formula for calculating Recall is:

$$Precision = \frac{TP}{TP + FN}$$

Where TP (True Positive) represents the number of samples correctly predicted as positive by the model; FP (False Positive) represents the number of samples incorrectly predicted as positive by the model; FN (False Negative)

| Experiment | Detail | | Dice | IoU | Precision | Recall | PA |
|---|---|---|---|---|---|---|---|
| Conventional Model | U-Net++ | | 0.6500 | 0.5100 | 0.7300 | 0.5900 | 0.9900 |
| | U-Net | | 0.5753 | 0.4700 | 0.6300 | 0.5300 | 0.9880 |
| | FCN-8s | | 0.0200 | 0.0105 | 0.0143 | 0.1246 | 0.8600 |
| | DeepLabV3+ | | 0.0600 | 0.0108 | 0.0198 | 0.1204 | 0.8700 |
| | GAN_DeepLabV3+ | | 0.0400 | 0.0106 | 0.0120 | 0.125 | 0.8500 |
| | GAN_FCN8s | | 0.0200 | 0.0134 | 0.0198 | 0.1202 | 0.8992 |
| | DynSegNet | | 0.6820 | 0.5602 | 0.7660 | 0.6525 | 0.9900 |
| Discriminator Comparison | ImageGAN | ResNet-18 | 0.5800 | 0.4435 | 0.6466 | 0.5906 | 0.9896 |
| | | VGG | 0.6300 | 0.5020 | 0.7331 | 0.5973 | 0.9902 |
| | | MobileNet | 0.6000 | 0.4653 | 0.6822 | 0.5888 | 0.9894 |
| | PatchGAN | ResNet-18 | 0.6400 | 0.5120 | 0.7240 | 0.6190 | 0.9900 |
| | | VGG | 0.6640 | 0.5395 | 0.7474 | 0.6640 | 0.9900 |
| | | MobileNet | 0.5800 | 0.4522 | 0.6968 | 0.5605 | 0.9886 |

Table 3 Performance Comparison of Baseline Models and Discriminator Architecture Variations

represents the number of samples incorrectly predicted as negative by the model.

PA (Pixel Accuracy) stands for the ratio of correctly classified pixels to the total number of pixels. PA intuitively reflects the model's classification accuracy at the pixel level. Although it can be influenced to some extent by background pixels, it remains a fundamental and important metric for evaluating the performance of semantic segmentation models. The formula for calculating PA is:

$$PA = \frac{TP + TN}{TP + TN + FP + FN}$$

Where TN (True Negative) represents the number of es correctly predicted as negative by the model.

### 4.3 Implementation Environment

The experimental environment was fully equipped with NVIDIA GeForce RTX 4090 GPUs, deployed on an Ubuntu 18.04 system. The optimizer setup involved 150 training epochs, configuring the discriminator and generator learning rates at 1e-4 and 7.5e-4 respectively. A consistent adversarial loss weight of 0.5 was utilized. Additionally, the training protocol incorporated a mechanism for dynamic learning rate adjustment mechanism (initial value: 0.01) and employed Sigmoid activation with a 0.5 threshold for output layer binarization. Batch size was fixed at 4 throughout all experiments.

## 5 Experimental Results

### 5.1 Backbone Network Evaluation

This study evaluated 10 mainstream backbone networks. As shown in Table 1, ResNet50 demonstrated superior comprehensive performance with a Dice coefficient of 0.682 and IoU of 0.5602, significantly outperforming other architectures. Notably, while AlexNet achieved the highest Precision of 0.7801, its limited Recall value (0.6136) constrained overall segmentation efficacy.

### 5.2 Conventional Model Comparison

Eight conventional models were systematically evaluated. The proposed method achieved a Recall value of 0.766, representing a 9.5% improvement over the optimal baseline (Unet: 0.6994) and 29.8% enhancement compared to Unet++ (0.59). With a Dice coefficient of 0.682, our approach outperformed traditional Unet (0.57) by 19.6%, exhibiting robust clinical applicability

| Experiment | Dice | IoU | Precision | Recall | PA |
| --- | --- | --- | --- | --- | --- |
| A+E | 0.5753 | 0.4746 | 0.6300 | 0.5311 | 0.9880 |
| A+B+E | 0.5927 | 0.4921 | 0.6521 | 0.5433 | 0.9917 |
| A+D+E | 0.6085 | 0.4953 | 0.6725 | 0.5557 | 0.9931 |
| A+B+C+E | 0.6254 | 0.5102 | 0.6897 | 0.5722 | 0.9935 |
| A+B+D+E | 0.6417 | 0.5373 | 0.7011 | 0.5917 | 0.9947 |
| A+B+C+D+E | 0.6676 | 0.5147 | 0.7121 | 0.6284 | 0.9949 |
| A+B+C+D+E+F | 0.682 | 0.5602 | 0.766 | 0.6525 | 0.9955 |

Table 4 Ablation Study Results of Progressive Module Integration on Retinal Vessel Segmentation Metrics

through its optimized balance between sensitivity and specificity.

### 5.3 Discriminator Architecture Analysis

Replacing PatchGAN with ImageGAN in GAN framework revealed critical insights. The VGG+PatchGAN configuration dominated three key metrics: Dice (0.664), IoU (0.5395), and Recall (0.664), surpassing ImageGAN counterparts by 5.4% in Dice and 7.5% in IoU. Similarly, ResNet18+PatchGAN excelled in four indicators - Dice (0.64), IoU (0.512), Recall (0.724), and Precision (0.619), demonstrating PatchGAN's superior capability in medical image feature discrimination.

### 5.4 Ablation Study

To validate the contributions of individual modules to model performance, seven controlled experiments were designed (as shown in the Table 4). The baseline model (A+E), comprising a U-shaped generator architecture and a PatchGAN discriminator framework, achieved a Dice coefficient of 0.5753 and a Recall value of 0.5311 in retinal vessel segmentation tasks, demonstrating the effectiveness of the foundational framework. Upon integrating residual blocks (B) into the A+B+E configuration, the Dice coefficient increased by 2.74% to 0.5927, with the IoU metric rising to 0.4921, indicating that residual connections effectively enhanced the transmission of deep features and alleviated gradient vanishing issues. Further incorporation of the attention mechanism (D) into the A+D+E architecture resulted in a significant 4.25 percentage point improvement in Precision to 0.6725 while maintaining a high pixel accuracy, confirming that the attention mechanism successfully focused on critical regions such as vessel boundaries.

By integrating the multi-scale feature extraction module ASPP (C), the A+B+C+E configuration achieved a Dice coefficient exceeding 0.625 and a Recall value of 0.5722, attributed to the adaptive modeling of varying vessel thicknesses through dilated convolution pyramids. When combining residual blocks and attention mechanisms in the A+B+D+E architecture, the model achieved a breakthrough in vascular detail capture, with the Dice coefficient reaching 0.6417 and IoU improving to 0.5373, demonstrating that multi-module collaboration significantly enhances feature representation. Finally, the complete model A+B+C+D+E+F, optimized by the dynamic adjustment network (F), attained optimal performance with a Dice coefficient of 0.682 and IoU of 0.5602. Its Precision of 0.766 highlights the architecture's ability to accurately distinguish vessels from background noise. Although pixel accuracy slightly decreased by 0.9 percentage points, all primary segmentation metrics showed substantial improvements, validating the efficacy of the integrated design.

### 5.5 Comparison with the SOTA methods

To quantitatively evaluate the advancement of our model, we compared the Dice coefficients of DynSegNet with mainstream SOTA methods under identical experimental conditions. As shown in Table 5, DynSegNet achieved a significantly superior Dice value of 0.682, demonstrating a 30.7% improvement over UNet3+ (0.5216), a 27.4% advantage against the Transformer-based SETR (0.5353), and an 11.8% enhancement compared to the second-best performer UFP (0.6098).

| Experiment | Dice |
| --- | --- |
| Unet3+ | 0.5216 |
| UFP | 0.6098 |
| SETR | 0.5353 |
| Swin | 0.5851 |
| **DynSegNet** | **0.6820** |

Table 5 Comparative performance of different methods

## 6 Conclusion

This study focuses on the segmentation task of hemorrhagic lesions in fundus images, proposing an innovative adversarial learning-based dynamic architecture adjustment method. The approach ingeniously integrates Hierarchical U-shaped encoder-decoder, residual blocks, attention mechanisms, and Atrous Spatial Pyramid Pooling (ASPP) modules, significantly improving the model's performance across multiple key evaluation metrics. Experimental results demonstrate that dynamically adjusting network architectures can flexibly adapt the weights of generator components and residual blocks according to adversarial loss, thereby enhancing the model's adaptability to lesions with varying morphologies and distributions, while improving segmentation accuracy and robustness. Furthermore, comparative experiments and ablation studies validate the model's effectiveness and reveal the critical contributions of each module to performance. Notably, the method exhibits substantial clinical application potential, enabling clinicians to formulate treatment plans more accurately, enhance therapeutic outcomes, and support early diagnosis and disease progression assessment in ophthalmology. However, the study also identified certain limitations, which will be discussed in detail in the following section.

## Limitation

First, the model's generalization capability requires further improvement to adapt to variations in imaging equipment, imaging conditions, and patient populations across different hospitals. Future research should involve validation on larger-scale and more diverse datasets. Second, the model's high computational complexity and extended training/inference time limit its efficiency in practical applications. Finally, there remains room for enhancement in data preprocessing and postprocessing procedures to accommodate images with varying resolutions and quality levels, while meeting diverse segmentation requirements. In future work, we will continue to optimize the model architecture to improve segmentation accuracy and efficiency, thereby providing more reliable technical support for the diagnosis and treatment of ophthalmic diseases.